\documentclass{article}
\usepackage{ijcai16}
\usepackage[fleqn]{amsmath}
\usepackage{times}
\usepackage{graphicx}
\usepackage{subcaption}
\title{UAV and Service Robot Coordination for Indoor Object Search Tasks}
\author{Sandeep Konam, Stephanie Rosenthal, Manuela Veloso \\ 
Carnegie Mellon University \\
sandeepkonam@cmu.edu, srosenthal@sei.cmu.edu, veloso@cmu.edu}

\begin{document}

\maketitle

\begin{abstract}
 	 Our CoBot robots have successfully performed a variety of service tasks in our multi-building environment including accompanying people to meetings and delivering objects to offices due to its navigation and localization capabilities. However, they lack the capability to visually search over desks and other confined locations for an object of interest. Conversely, an inexpensive GPS-denied quadcopter platform such as the Parrot ARDrone 2.0 could perform this object search task if it had access to reasonable localization. In this paper, we propose the concept of coordination between CoBot and the Parrot ARDrone 2.0 to perform service-based object search tasks, in which CoBot localizes and navigates to the general search areas carrying the ARDrone and the ARDrone searches locally for objects. We propose a vision-based moving target navigation algorithm that enables the ARDrone to localize with respect to CoBot, search for objects, and return to the CoBot for future searches. We demonstrate our algorithm in indoor environments on several search trajectories. 
\end{abstract}

\section{Introduction}

Autonomous mobile service robots are being sought after for deployment in homes, offices, hospitals, warehouses and restaurants for searching, handling, picking and delivering goods/objects. For example, our CoBot service robots have been deployed continuously in our multi-building environment for many years performing tasks such as delivering messages, accompanying people to meetings, and transporting objects to offices in our environment \cite{6386300}. CoBots autonomously localize and navigate in our office environment, while effectively avoiding obstacles using a modest variety of sensing and computing devices, including a vision camera, a Kinect depth-camera, a small Hokuyo LIDAR, a touch-screen tablet, microphones and speakers, as well as wireless communication \cite{5509842}. 

However, our CoBots do not have capability to navigate small spaces to visualize the objects on desks and other tall surfaces, and as a result they fail to efficiently search for objects of interest without human assistance. We propose multi-robot coordination for service robots to utilize and build upon the capabilities of our CoBots \cite{6386300} by combining the complementary strengths of another autonomous robot platform. The combination of autonomous platforms will allow the robots to perform more service tasks effectively and without human intervention. 

\begin{figure}[t]
\begin{center}
  \includegraphics[width=35mm]{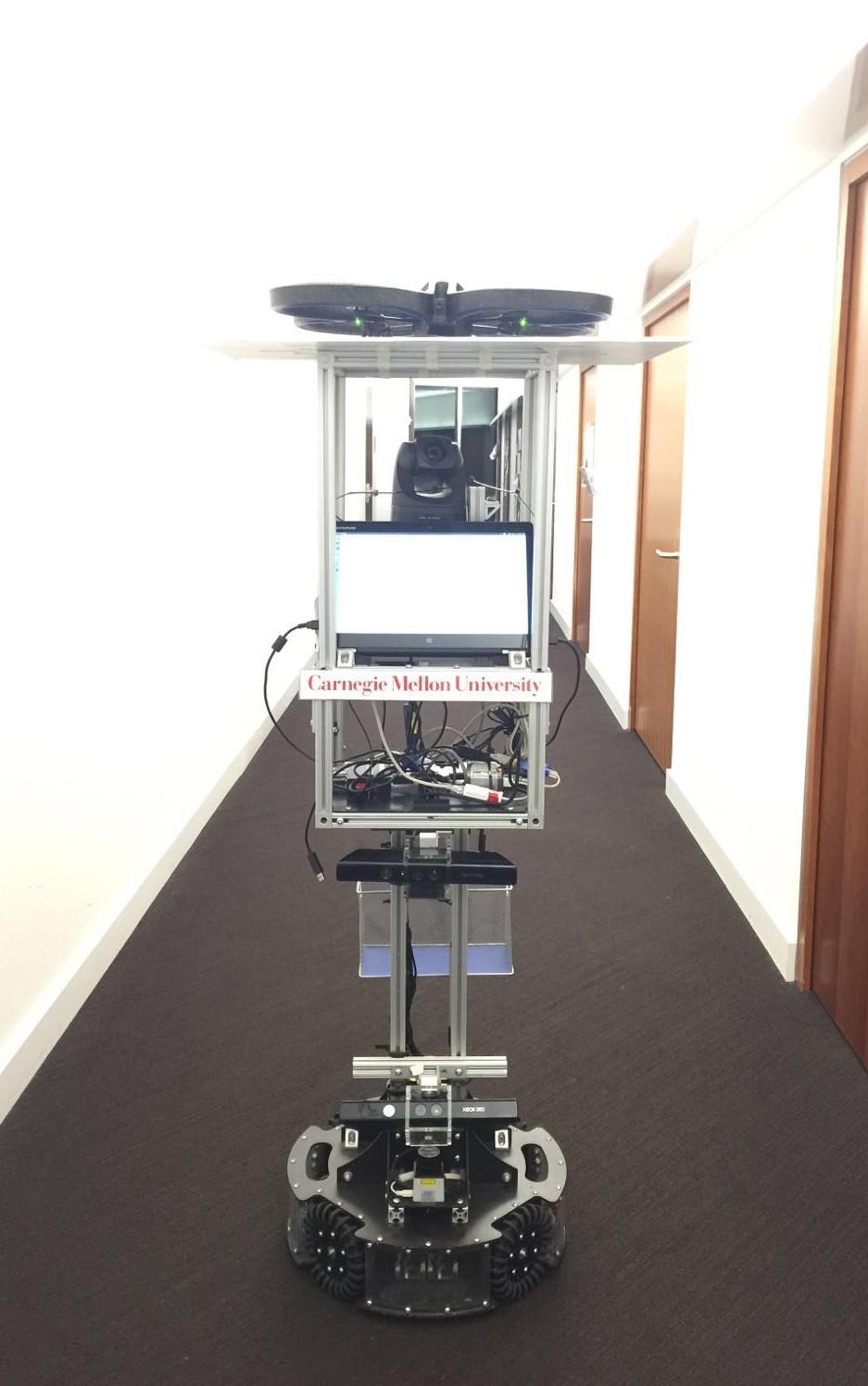}
\end{center}
  \caption{CoBot carrying ARDrone.}
  \label{cobotdrone} 
\end{figure}

In this work, we focus our study on combining CoBot with the off-the-shelf Parrot ARDrone 2.0 Unmanned Aerial Vehicle (UAV) to perform service-based object search tasks by taking off from and landing on CoBot between searches (Figure~\ref{cobotdrone}). Though the ARDrone was primarily launched as a consumer electronic device, it was adopted by universities and research institutions as a research platform \cite{Krajník2011}. While it can be controlled autonomously, hover above the environment, and its two cameras allow it to see objects from a variety of angles, it lacks the endurance, payload, power efficiency, and localization and navigation capabilities that the CoBots have. By allowing the ARDrone to take off from and land on CoBot, we leverage the robust localization and navigation capabilities of CoBot with ARDrone's capability to maneuver easily through indoor environments and search for object of interest. CoBot autonomously carries the ARDrone across corridors in our environment to designated search spaces where the drone can then perform search operations and return its results to CoBot. 

Given these new capabilities, the ARDrone must still be able to navigate locally to and from CoBot. We propose a vision-based moving target approach for the navigation of ARDrone in which the drone uses a camera-based coordinate system to track the direction the robot and object has moved, find and hover above objects, and reverse those trajectories to return to its starting location under uncertainty. This approach requires little computation and yet allows the drone to perform its search task from any location in any environment. We demonstrate that the algorithm efficiently finds objects in our environment during search tasks and can return to its starting location after it is finished searching.

\section{Related Work}
Given the reliance on our CoBot robots for most localization and navigation, the challenge of our multi-robot coordination task is the localization and navigation of our drone. Drones or aerial vehicles typically use IMU and GPS data fused together to estimate its state based on which appropriate navigation algorithm is used. However, in many indoor conditions, the strength of GPS signal might not be enough to enable precise navigation. Cameras offer an inexpensive and reliable solution using a variety of ``vision-servoing'' schemes that rely on visual features as feedback signals \cite{visserv}. There has been immense research in this area, leading to several categorizations of the approaches. 

The primary distinction between visual-servoing techniques whether they use additional sensor data to fuse with the visual information  \cite{chatterji1997gps,roberts2002low} or only use visual information \cite{zhang1999visual}. 
Among techniques that rely only on images, there have been approaches that require a priori knowledge of landmarks \cite{sharp2001vision,jones2006vision} and those that do not require any priori information about landmarks \cite{shakernia1999landing,yakimenko2002unmanned,koch2006vision}. Distinction can also be made according to the type of computer vision techniques employed to extract information such as stereo vision \cite{trisiripisal2006stereo}, optic flow \cite{barber2007autonomous}, epipolar geometry \cite{shakernia1999landing,wu2005vision,webb2007vision,caballero2006improving} and simultaneous localization and mapping (SLAM) \cite{bryson2008observability,kim2004autonomous}. 

Most of the existing techniques are computationally expensive and precision is often traded with real-time computation on small platforms such as UAVs. The ARDrone was previously used for visual SLAM based navigation \cite{engel2012camera}, autonomous navigation of hallways and stairs  \cite{bills2011autonomous} and reactive avoidance in natural environments \cite{ross2013learning}. Inferring 3D structure from multiple 2D images is challenging because aerial vehicles are not equipped with reliable odometry and building a 3D model is computationally very expensive. \cite{bills2011autonomous} instead compute perspective cues to infer about 3D environment. Unfortunately most indoor environments don't possess distinct corner-type features to provide the desired perspective cues.

Inspired by the accuracy of visual tracking methods, we designed \textit{vision-based moving target navigation algorithm} for drones that does not rely on any environment features and is computationally minimal. In our approach, if an object is visible in the robot's visible range, the robot aims to minimize the distance between itself and the center of the object. If no object is visible, the algorithm ``imagines'' where one could be outside of its visible range and then navigates to it in a similar way. Using the knowledge of the robot's current speed and travel time, the robot navigates in any arbitrary search pattern by computing where imaginary objects should be placed. We next discuss the multi-robot coordination object search task and then describe our navigation algorithm for completing the service task in detail.


\section{Multi-Robot Coordination for Object Search Tasks}

Our CoBot service robots have been deployed in our environment for many years performing tasks such as delivering messages, accompanying people to meetings, and transporting objects to offices \cite{6386300}. However, our CoBots do not have capability to navigate confined spaces such as offices to visualize the objects in them, and as a result they require human assistance to find required objects in search tasks. We propose a multi-robot coordination in which a second platform - the Parrot ARDrone - can perform the object search task for CoBot, while at the same time relying on CoBot for its localization and navigation. We describe the robot's capabilities and the joint task before focusing on the remaining challenge of drone localization and navigation.

\subsection{CoBot Capabilities}

CoBot is a four-wheeled omni-directional robot, equipped with a short-range laser range-finder sensor and a depth camera for sensing (Figure~\ref{cobotdrone}). An on-board tablet provides significant computation for the robot to localize and navigate autonomously, as well as a method of communication with humans in the environment. The CoBot robots can perform multiple classes of tasks, as requested by users through a website \cite{ventura2013web}, in person through speech \cite{6631186} or through the robot's touch screen. All tasks can be represented as pick up and delivery tasks of objects or people. Because CoBot is not able to navigate in confined spaces safely, it lacks the ability to visually search for objects to pick up and drop off, instead relying on humans. 
	
\subsection{ARDrone Capabilities}

Parrot ARDrone 2.0 has a 3-axis accelerometer, 3-axis gyroscope, Pressure sensor, Ultrasonic sensors, Front camera and a Bottom/Vertical camera. With complete reliance on WiFi to transfer all of its sensor data onto a larger computational platform, the drone cannot employ typical computationally-expensive localization and navigation methods. It can however send its video feed for simple object detection within the local camera frame (e.g., color thresholding for colored marker detection). Velocity commands are sent back to the drone for vision-based navigation.

\begin{figure*}[t!]
    \centering
    \begin{subfigure}[t]{0.5\textwidth}
        \centering
         \includegraphics[width=55mm,height=35mm]{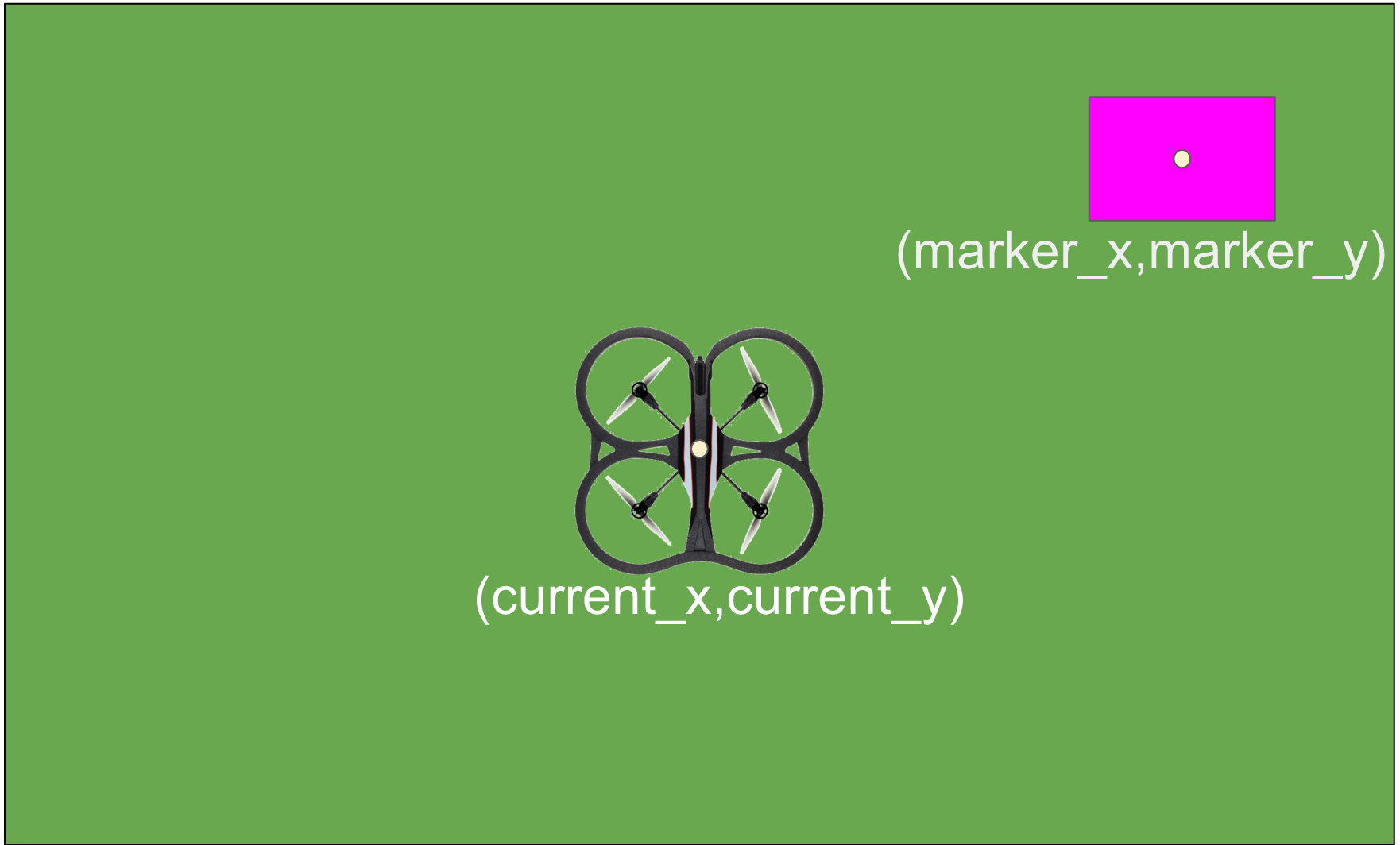}
        \caption{Pink marker as seen through bottom camera.}
        \label{imagecoordinates}
    \end{subfigure}%
     ~ 
    \begin{subfigure}[t]{0.5\textwidth}
        \centering
        \includegraphics[width=55mm,height=35mm]{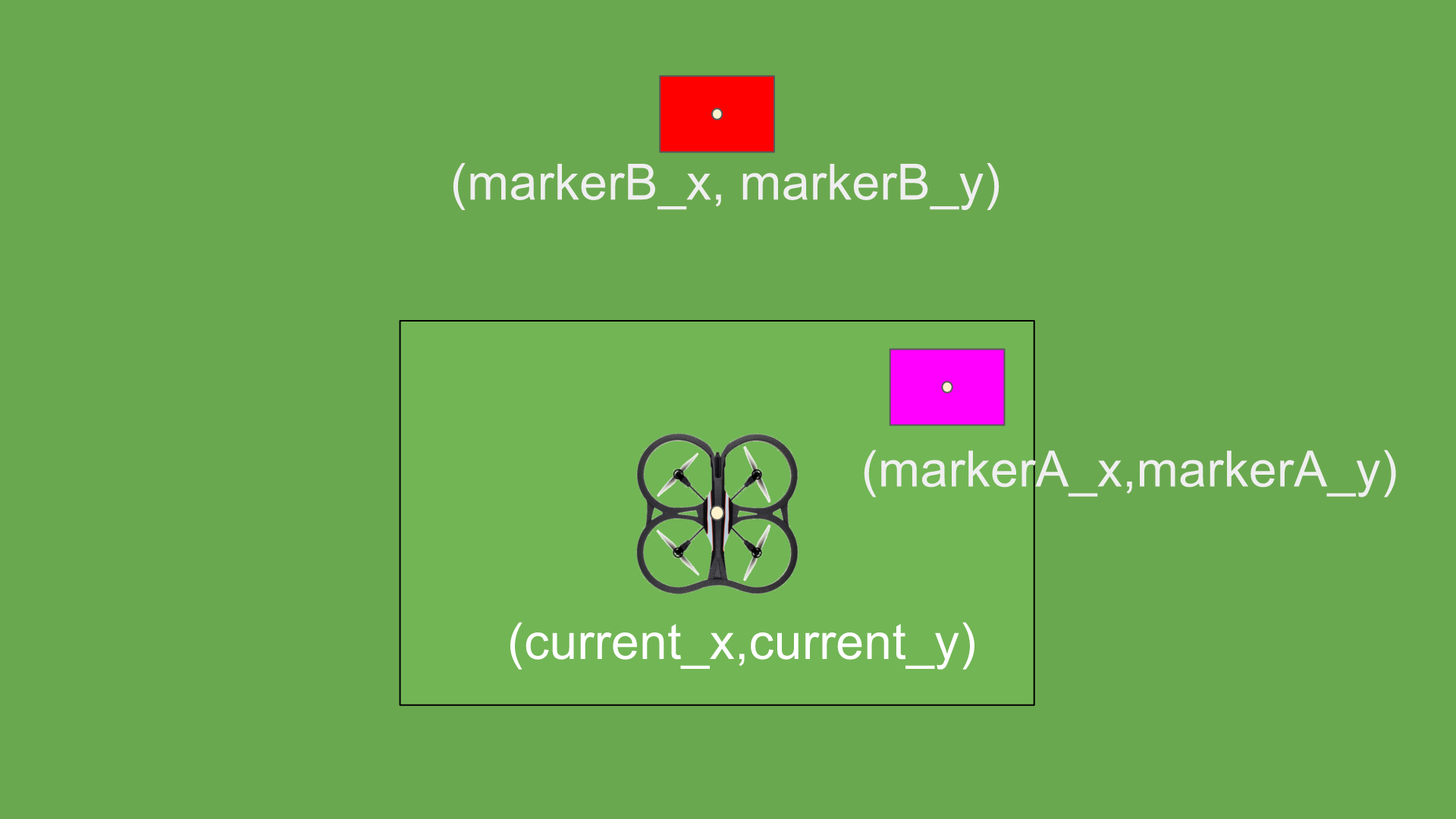}
         \caption{The black rectangle represents the image captured by bottom camera of drone. One of the colored markers (pink marker) is in the field of view where as the other (red marker) can be reached through forward search.} 
          \label{beyondimage}
    \end{subfigure}
    ~ 
    \begin{subfigure}[t]{0.5\textwidth}
        \centering
        \includegraphics[width=55mm]{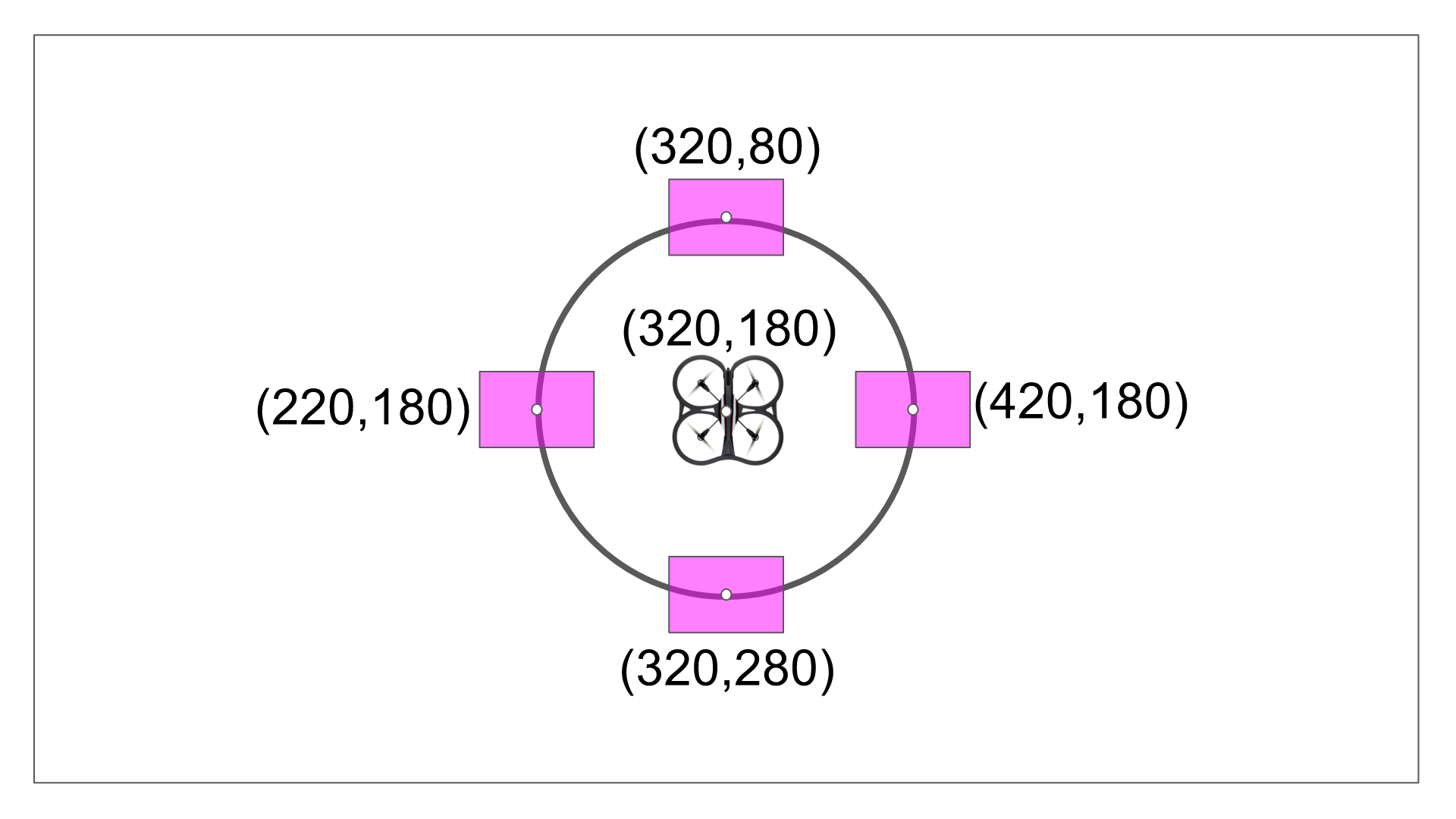}
         \caption{Sample Coordinates corresponding to the four primary directions and center of the drone marked on a 640x360 image.}
           \label{samplecoordinates}
    \end{subfigure}
 
    \caption{}
\end{figure*}

\subsection{Multi-Robot Coordination Task}

Considering the above mentioned capabilities of CoBot and ARDrone, we notice that the ARDrone can perform object search tasks with small computational loads but without reliable localization while CoBot can accurately localize and navigate in its safe regions. We propose coordination between them to effectively search for an object of interest in an indoor environment. In particular, CoBot carries the drone to a region of interest to search and then the drone can search locally by tracking its relative motion trajectory after taking off from CoBot. After it finishes its search, it can reverse the trajectory or perform another search to land on CoBot and move to another location.

For the search task to be performed efficiently, the drone should be able to navigate in its local coordinate space. Lack of reliable camera-based localization algorithms for the resource constrained ARDrone forced us to opt for visual-servoing techniques. While performing the search task, the image provided by bottom camera of drone can only provide information about presence or absence of the marker being searched for, but doesn't provide any cues that facilitate search. We contribute our vision-based moving target navigation algorithm to overcome the challenges of localizing and navigating without any visual cues.

\section{Vision-Based Moving Target Navigation}

In moving target navigation algorithms, a robot continuously aims to minimize the distance from its current location and another target point in its coordinate space. Maintaining the target in the same place over time allows the robot can navigate directly to it (i.e., for hovering over an object of interest). By moving the target point in a trajectory at a constant velocity, the robot follows the same trajectory. We use this moving target navigation algorithm in order for the ARDrone to search the environment and track a marker when it finds one, noting that the challenge of this algorithm is determining the local coordinate space to move the point in. We next describe our coordinate space for the moving target algorithm. 


\subsection{Image Coordinate System}

Because the ARDrone's only frame of reference is its bottom camera, we use this camera's coordinate frame as the local frame of reference. The ARDrone's image is represented as pixels in its bottom camera time (640x360 in this work). When a marker is found in an image (Figure~\ref{imagecoordinates}), the computational platform sends velocity commands proportional to the distance from the center of the robot (image) to the center of the target marker:
\begin{equation}
error_x = (target_x - current_x) 
\end{equation}
\begin{equation}
\begin{aligned}
error_y = (target_y - current_y) 
 \end{aligned}
\end{equation}
\begin{equation}
vel_x=k*(error_y) 
\end{equation}
\begin{equation}
vel_y=k*(error_x) 
\end{equation}

In the above equations, $error_x$ refers to the difference in x-coordinates of the UAV'™s current position ($current_x$) and that of virtual marker'€™s center ($target_x$). Similar notation applies to $error_y$. $vel_x$ refers to the linear velocity in x-direction and $vel_y$ refers to the linear velocity in y-direction. Proportional (P) controller is deployed where $k$ is empirically tuned to be 0.0005.	

Figure~\ref{samplecoordinates} demonstrates markers in the 4 cardinal directions of the robot's image frame. Since the marker exists in the image coordinate system, the error between the center of marker and the position of UAV tends to reduce as the UAV approaches the marker. Since the velocity is directly proportional to the error, the UAV starts hovering once the error becomes zero or falls within a certain threshold.

\begin{figure*}[t!]
    \centering
    \begin{subfigure}[t]{0.5\textwidth}
        \centering
         \includegraphics[width=55mm,height=35mm]{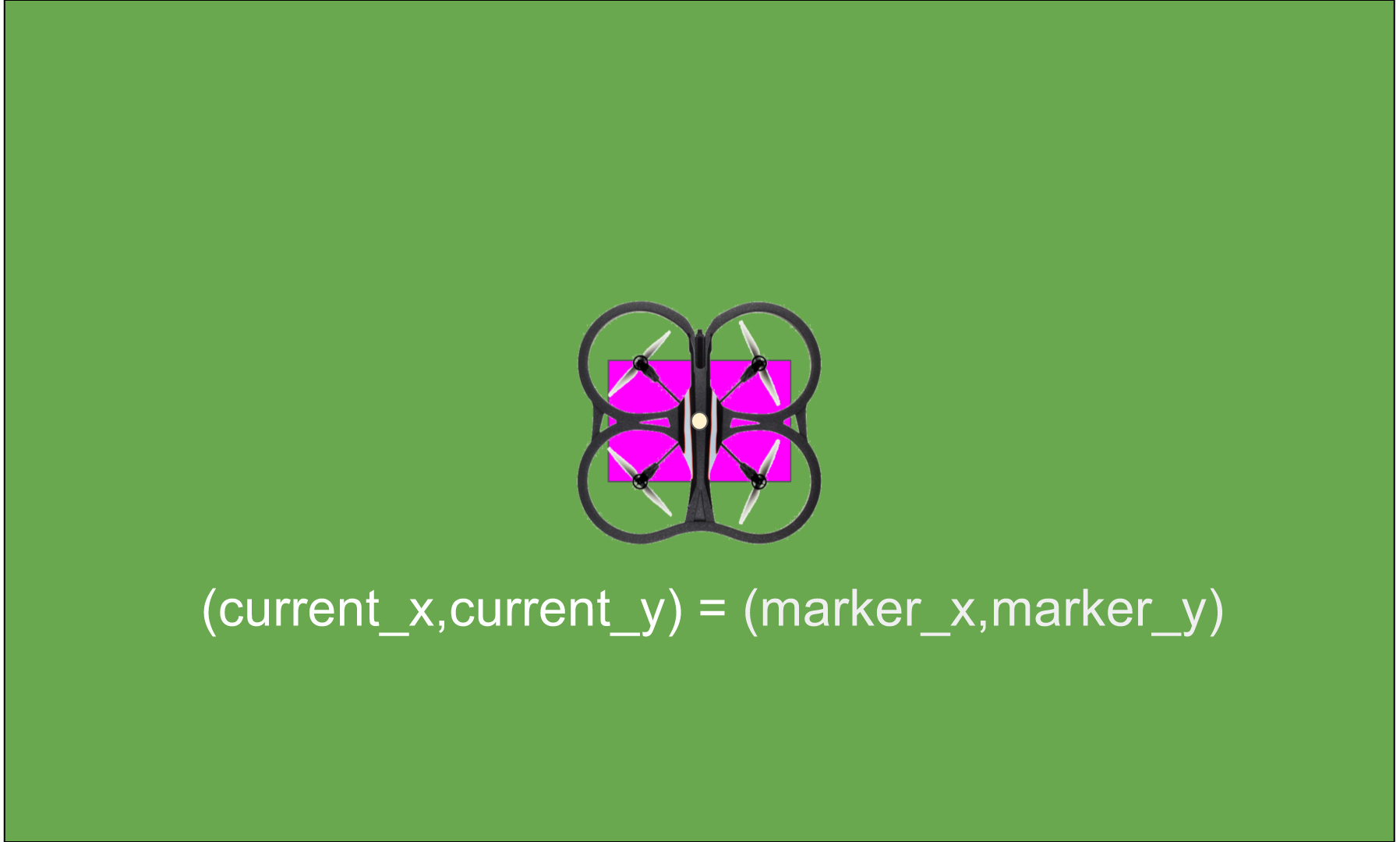}
        \caption{Through error minimization, UAV has reached center of the marker and hovers above it.}
         \label{3a}
    \end{subfigure}%
     ~ 
    \begin{subfigure}[t]{0.5\textwidth}
        \centering
        \includegraphics[width=55mm,height=35mm]{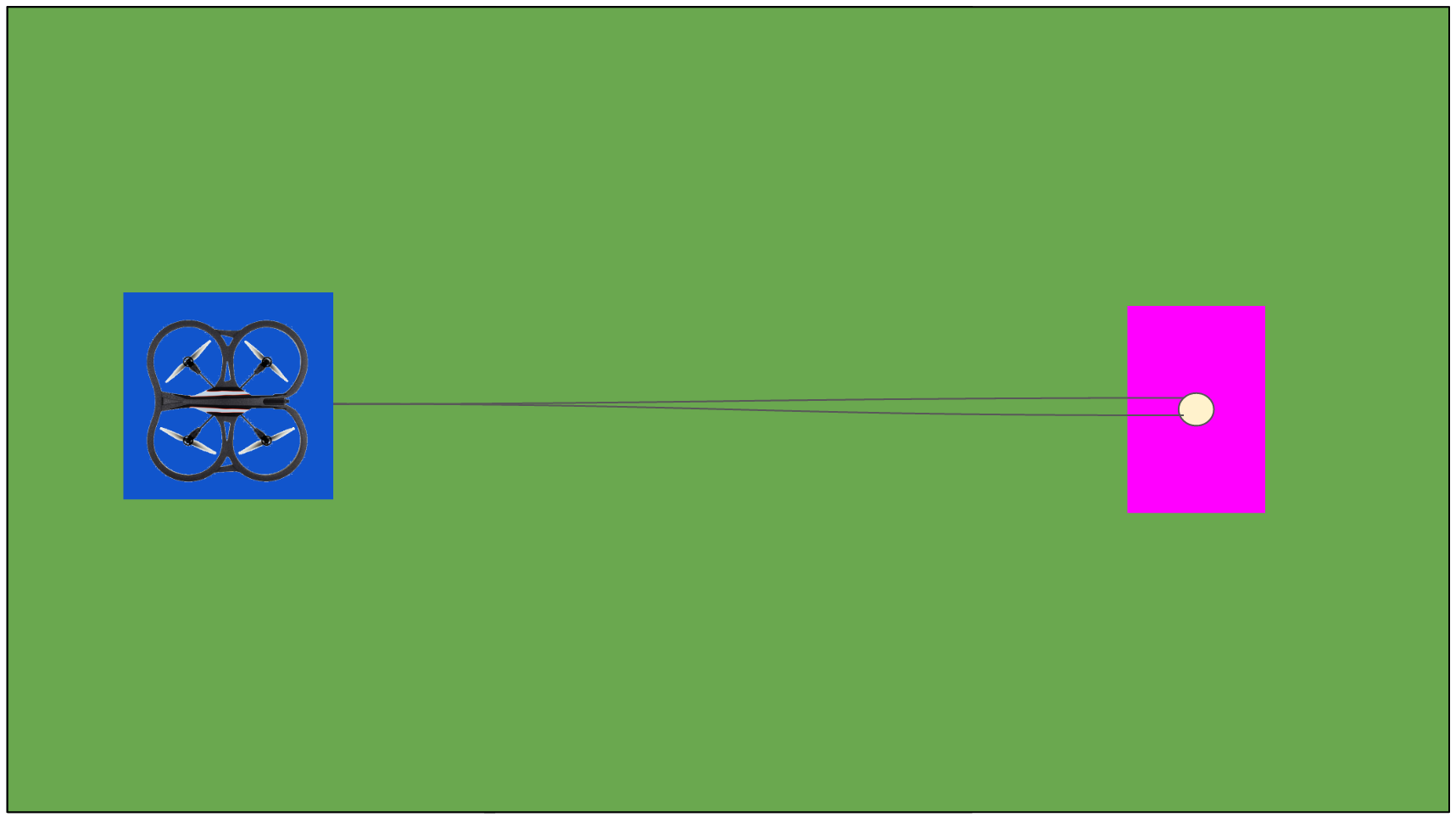}
          \caption{UAV landing on CoBot after performing forward search and detecting the marker.}
           \label{3b}
    \end{subfigure}
      ~ 
    \begin{subfigure}[t]{0.5\textwidth}
        \centering
       \includegraphics[width=30mm,scale=1]{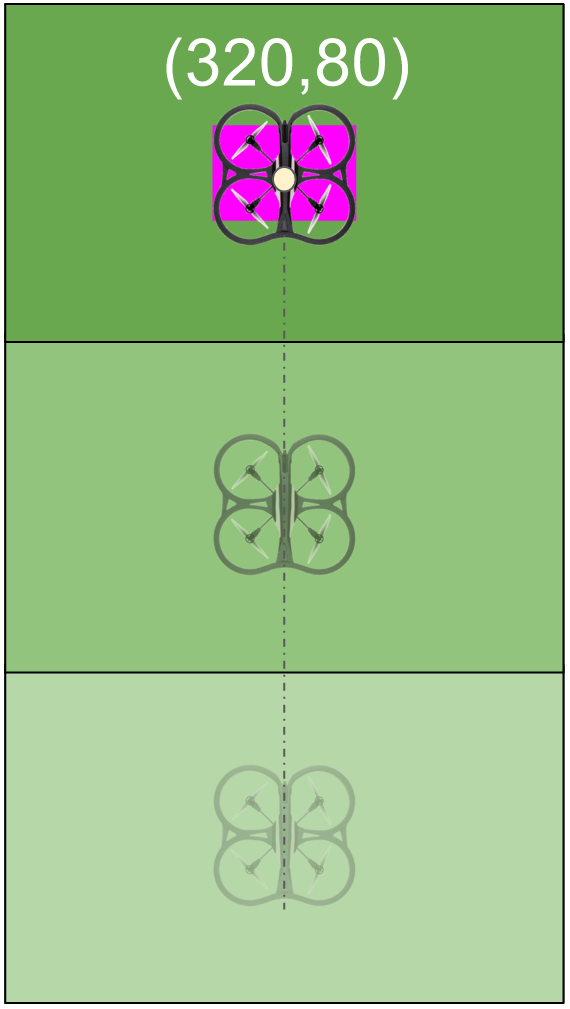}
           \caption{UAV navigating forward searching for the marker. Above three parts include sequential captures from the bottom camera of UAV.}
            \label{3c}
    \end{subfigure}
   \caption{}
\end{figure*}

\subsection{Imagining Beyond the Image Frame}

When the target marker is not in the image frame, we would like the robot to search for it using moving target navigation. By referring to pixel locations outside of the image frame (Figure~\ref{beyondimage}), the robot imagines where the target should be and navigates towards it. Imagined coordinates could either be random or based on knowledge of direction in which object went out of view or generated corresponding to a trajectory. In the current work, we generate coordinates representing imagined markers based on the required trajectory robot has to follow. However since it is an imagined marker at a constant distance from the current position of the robot, the error remains constant and the robot maintains a constant velocity using the equations 1-4. For instance, if we want the robot to perform search operation in a square path, we generate coordinates corresponding to imaginary markers at the four corners of square. By creating trajectories of imagined markers outside (or even inside) of the image area, the robot navigates using those coordinates and eventually detects object of interest.

\subsection{Moving Target Navigation Algorithms}

In order to demonstrate the applicability of the moving target algorithm, we designed several search trajectories as finite state machines in addition to hovering behavior over a marker in the vision frame. We describe each in turn. Then, we will show results for experiments for each algorithm.

\subsubsection{Navigation to marker in vision frame}

Navigation to a marker detected in the field of view of the UAV relies entirely on the image coordinate system. After the marker is detected, its center becomes the target coordinate for the UAV. UAV navigates towards the center of the marker with velocity proportional to the error between its current position and the target coordinate.

\begin{figure*}[t!]
    \centering
    \begin{subfigure}[t]{0.5\textwidth}
        \centering
         \includegraphics[width=65mm]{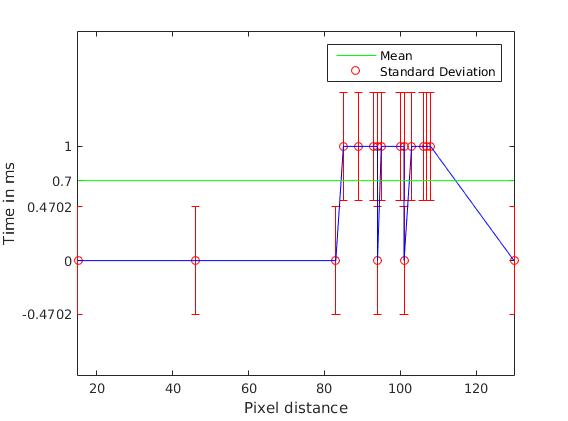}
         \caption{Task1: Navigation to marker in vision frame.}
          \label{t1}
    \end{subfigure}%
     ~ 
    \begin{subfigure}[t]{0.5\textwidth}
        \centering
	\includegraphics[width=65mm]{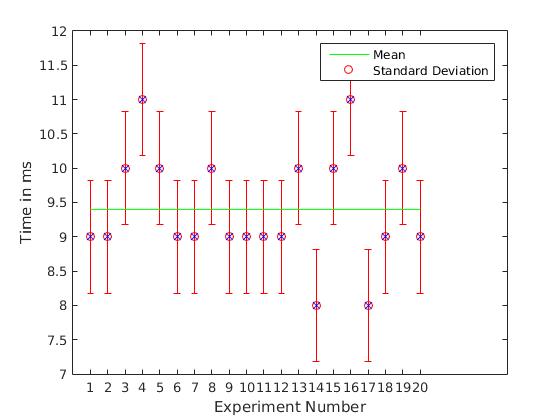}
	\caption{Task2: Forward Search and hover above marker.}
	 \label{t2}
    \end{subfigure}
      ~ 
    \begin{subfigure}[t]{0.5\textwidth}
        \centering
	  \includegraphics[width=65mm]{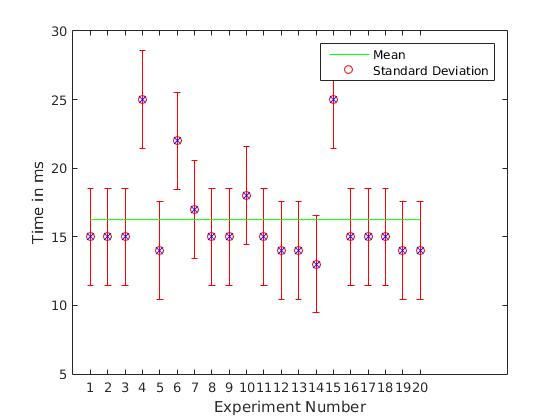}
	  \caption{Task3: Forward Search, Returning Home, Landing.}
	   \label{t3}
    \end{subfigure}
   \caption{Plots indicating standard deviation and mean of the observed data for different tasks}
\end{figure*}

\subsection{Forward Search and hover above marker}

As depicted in Figure~\ref{samplecoordinates}, (320,80) refers to the center of imagined marker corresponding to forward direction.  As the UAV navigates forward, the video feed from the bottom camera is used to search for the marker. Since the marker is distinctively colored from the background, thresholding algorithm suffices to let the UAV know if it sees the marker in its field of view. Once the marker is detected, UAV navigates towards the marker to hover over it.

\subsection{Forward Search, Returning Home, Landing}

Forward search is performed as mentioned previously. After detecting the marker, UAV starts navigating backward to the home (place where it started search from) by reversing the trajectory of imagined target points. Note that in this case, reversing the trajectory is equivalent to the forward search problem where home is another distinctively colored marker. After detecting the marker corresponding to home, drone navigates to the center of marker using the image coordinate system. It lands on the home marker after centering itself using error-minimization between the center of marker and its own position.

\subsection{Coordination between CoBot and UAV to perform visual search}

In this task, CoBot carries UAV down the hallway to the destined search area. UAV performs forward search, returns home and lands on CoBot. As shown in Figure~\ref{3b}, the blue marker represents the CoBot. The ARDrone performs two search operations: one for the search object - pink marker and other for the CoBot - blue marker. 

\begin{figure*}[t]
\centering
  \includegraphics[width=\linewidth]{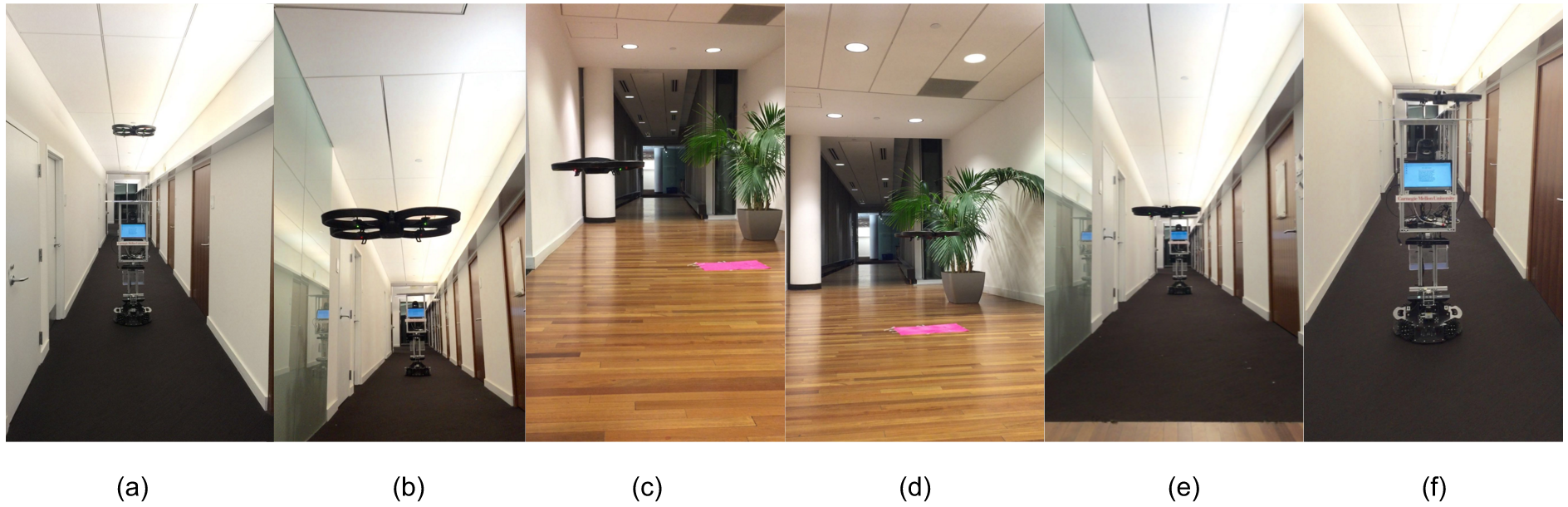}
  \caption{Coordination between CoBot and UAV to perform visual search task.(a) Drone taking off from the CoBot, (b)-(c) Drone performing forward search for target marker, (d) Drone hovering above marker after reaching it, (e) Drone performing backward search for CoBot, (f) Drone landing above CoBot.}
   \label{all}
\end{figure*}

\section{Experimental Results}

We evaluated the performance of our proposed vision-based moving target
navigation algorithm on ARDrone 2.0. through a series of experiments presented in this section. ARDrone has 1GHz ARMv7 Processor rev 2 (v7l) with 1Gbit DDR2 RAM and a VGA downward camera of 360p resolution at 30fps. Images captured are of 640x360 resolution. It is controlled via Wi-Fi through a laptop with Intel® Core™ i7-6700HQ CPU and 16 GB RAM. 

Each of the experiments presented in this section correspond to the tasks detailed in Section 4. We have performed each experiment 20 times to test the consistency of the proposed algorithm. Mean and standard deviation is calculated for each of the experiments and is depicted in the charts. Results suggest that the proposed algorithm is reliable for search tasks. 

\subsection{Navigation to marker in vision frame}

In this experiment, drone navigates to the center of the detected marker. Though drone takes off from the same place in all experiments, due to the unavoidable drift associated with the drone while taking off, pixel distance between the center of marker and the position of drone (center of the image) changes as represented on the x-axis of plot shown in the Figure~\ref{t1}. Due to the precision of timestamp being 1ms, it is possible that 0.9ms is considered as 0ms. 14 out of 20 times, drone takes 1 ms and 6 out of 20 times, it takes 0 ms to reach the center of marker. As depicted in the Figure~\ref{t1}, mean is around 0.7 and standard deviation is 0.4702.

\subsection{Forward Search and hover above marker}

In this experiment, a marker is placed 2.0 meters infront of the drone. Drone performs forward search and hovers above the marker once it reaches within a threshold of 50 units of pixel distance from the center of marker. Each experiment is performed 20 times. Results show standard deviation of 0.8208 with a mean of 9.40 as depicted in Figure~\ref{t2}.

\subsection{Forward Search, Returning Home, Landing}

\begin{figure}[h]
\begin{center}
  \includegraphics[width=60mm]{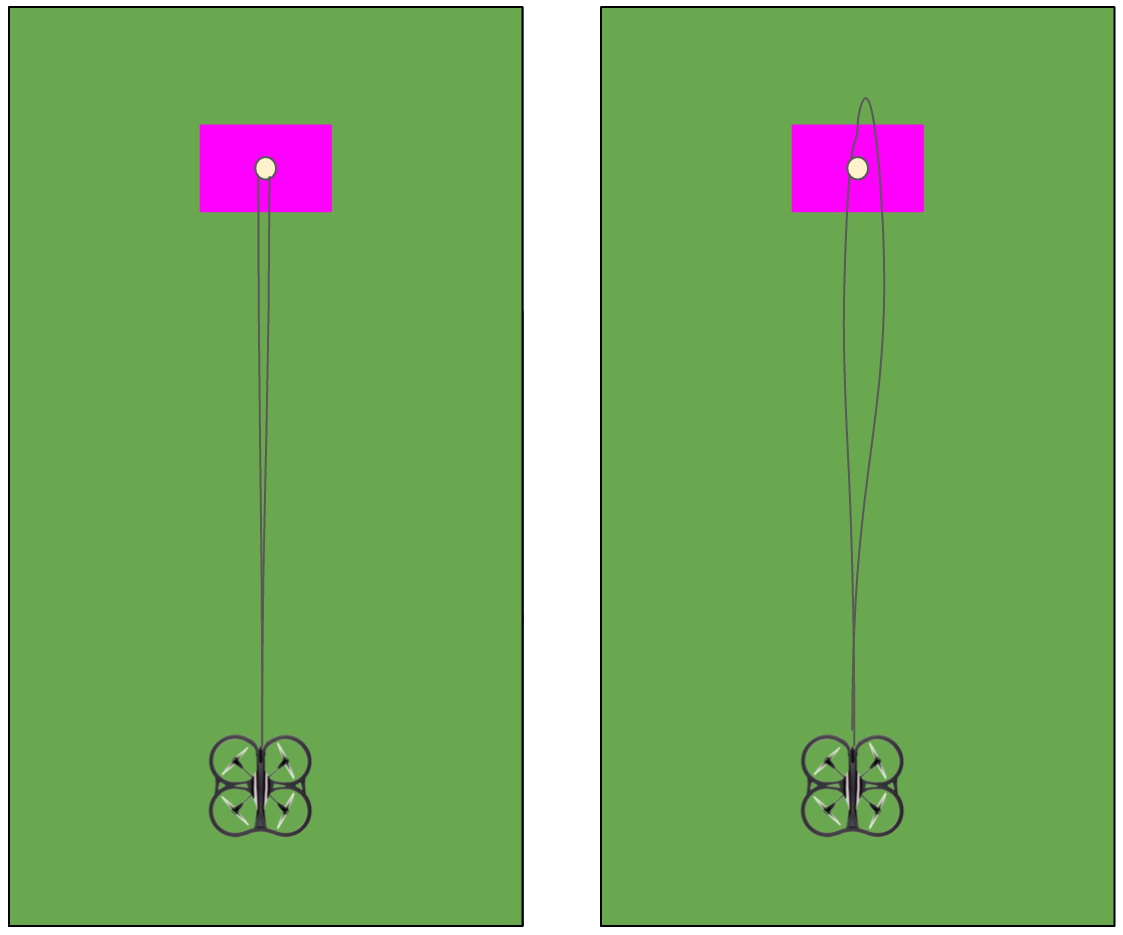}
  \caption{Approximate sketch of the path traversed by drone in two scenarios: left part shows unaffected navigation during forward and backward motion, where as right part shows the effect of drift on the course of navigation.}
   \label{last}
\end{center}
\end{figure}

In this experiment, drone performs forward search and returns back to the home after detecting the marker which is placed 2.0 meters infront of the drone. Over 20 experiments, results show a standard deviation of 3.5522 and mean of 16.2500 as depicted in the Figure~\ref{t3}. For 17 times, time reported is similar with a standard deviation of 1.1663, but there are two cases where the time taken is 25 ms and one case where it is 22 ms, including which increased standard deviation to 3.5522. Through the data log, it is observed that in those three cases, drone has drifted after detecting the marker and required additional time for marching backwards to the home. Left part of the Figure~\ref{last} shows approximate sketch of the path followed by the drone in cases where similar time was consumed, where right part shows path that looks more widened near the marker due to the drift of the drone.

\subsection{Coordination between CoBot and UAV to perform visual search task}

We bring together all the above building blocks and accomplish the visual search task by establishing coordination between CoBot and UAV. It can be observed through series of images illustrated in Figure 5. CoBot carries the drone using its localization and navigation capabilities to the desired search area. Drone takes off from the CoBot as shown in Figure 5a and starts navigating forward as shown in Figure 5b.  As it navigates forward, drone simultaneously searches for the pink marker as shown in Figure 5c. Once it locates the marker in its field of view, drone navigates to the center of marker through error minimization. Figure 5d shows the drone hovering above the marker. After reaching the center of marker, drone starts navigating backwards to the CoBot as shown in Figure 5e. As it navigates backwards, it keeps searching for the blue marker indicative of the CoBot. Once the blue marker is detected, drone lands on the CoBot as shown in Figure 5f.

\section{Conclusion}

In this paper, we presented coordination between CoBot and drone leveraging the robust localization and navigation capabilities of CoBot with ARDrone's capability to maneuver easily through indoor environments and search for an object of interest. As to enable reliable navigation of drone to perform search task, we proposed a vision-based moving target navigation algorithm. We evaluated the proposed system on several search trajectories. Observed results affirm that the proposed algorithm could be used effectively for the service search tasks. Aerial vehicles are one of the challenging and complex control systems easily affected by external factors such as wind and internal factors such as system stabilization. Improving stability of the drone could help advance several other service tasks apart from search.

\bibliographystyle{ijcai16}
\bibliography{ijcai16}

\begin{thebibliography}{}

\bibitem[\protect\citeauthoryear{Barber \bgroup \em et al.\egroup
  }{2007}]{barber2007autonomous}
D~Blake Barber, Stephen~R Griffiths, Timothy~W McLain, and Randal~W Beard.
\newblock Autonomous landing of miniature aerial vehicles.
\newblock {\em Journal of Aerospace Computing, Information, and Communication},
  4(5):770--784, 2007.

\bibitem[\protect\citeauthoryear{Bills \bgroup \em et al.\egroup
  }{2011}]{bills2011autonomous}
Cooper Bills, Joyce Chen, and Ashutosh Saxena.
\newblock Autonomous mav flight in indoor environments using single image
  perspective cues.
\newblock In {\em Proceedings of ICRA 2011}, pages 5776--5783. IEEE, 2011.

\bibitem[\protect\citeauthoryear{Biswas and Veloso}{2010}]{5509842}
J.~Biswas and M.~Veloso.
\newblock Wifi localization and navigation for autonomous indoor mobile robots.
\newblock In {\em Proceedings of ICRA 2010}, pages 4379--4384, May 2010.

\bibitem[\protect\citeauthoryear{Bryson and
  Sukkarieh}{2008}]{bryson2008observability}
Mitch Bryson and Salah Sukkarieh.
\newblock Observability analysis and active control for airborne slam.
\newblock {\em Aerospace and Electronic Systems, IEEE Transactions on},
  44(1):261--280, 2008.

\bibitem[\protect\citeauthoryear{Caballero \bgroup \em et al.\egroup
  }{2006}]{caballero2006improving}
Fernando Caballero, Luis Merino, Joaquin Ferruz, and Anibal Ollero.
\newblock Improving vision-based planar motion estimation for unmanned aerial
  vehicles through online mosaicing.
\newblock In {\em ICRA 2006}, pages 2860--2865. IEEE, 2006.

\bibitem[\protect\citeauthoryear{Chatterji \bgroup \em et al.\egroup
  }{1997}]{chatterji1997gps}
GB~Chatterji, PK~Menon, and B~Sridhar.
\newblock Gps/machine vision navigation system for aircraft.
\newblock {\em Aerospace and Electronic Systems, IEEE Transactions on},
  33(3):1012--1025, 1997.

\bibitem[\protect\citeauthoryear{Engel \bgroup \em et al.\egroup
  }{2012}]{engel2012camera}
Jakob Engel, J{\"u}rgen Sturm, and Daniel Cremers.
\newblock Camera-based navigation of a low-cost quadrocopter.
\newblock In {\em Proceedings of IROS 2012}, pages 2815--2821. IEEE, 2012.

\bibitem[\protect\citeauthoryear{Hutchinson \bgroup \em et al.\egroup
  }{1996}]{visserv}
Seth Hutchinson, Gregory~D Hager, and Peter~I Corke.
\newblock A tutorial on visual servo control.
\newblock {\em Robotics and Automation, IEEE Transactions on}, 12(5):651--670,
  1996.

\bibitem[\protect\citeauthoryear{Jones \bgroup \em et al.\egroup
  }{2006}]{jones2006vision}
Christopher~G Jones, JF~Heyder-Bruckner, Thomas~S Richardson, and DC~Jones.
\newblock Vision-based control for unmanned rotorcraft.
\newblock In {\em In Proceedings of the AIAA Guidance, Navigation, and Control
  Conference, AIAA-2006-6684, Keystone, CO.}, 2006.

\bibitem[\protect\citeauthoryear{Kim and Sukkarieh}{2004}]{kim2004autonomous}
Jonghyuk Kim and Salah Sukkarieh.
\newblock Autonomous airborne navigation in unknown terrain environments.
\newblock {\em Aerospace and Electronic Systems, IEEE Transactions on},
  40(3):1031--1045, 2004.

\bibitem[\protect\citeauthoryear{Koch \bgroup \em et al.\egroup
  }{2006}]{koch2006vision}
Andreas Koch, Hauke Wittich, and Frank Thielecke.
\newblock A vision-based navigation algorithm for a vtol-uav.
\newblock In {\em In AIAA Guidance, Navigation and Control Conference and
  Exhibit (Vol. 6, p. 11).}, 2006.

\bibitem[\protect\citeauthoryear{Kollar \bgroup \em et al.\egroup
  }{2013}]{6631186}
T.~Kollar, V.~Perera, D.~Nardi, and M.~Veloso.
\newblock Learning environmental knowledge from task-based human-robot dialog.
\newblock In {\em ICRA 2013}, pages 4304--4309, May 2013.

\bibitem[\protect\citeauthoryear{Krajn{\'i}k \bgroup \em et al.\egroup
  }{2011}]{Krajník2011}
Tom{\'a}{\v{s}} Krajn{\'i}k, Vojt{\v{e}}ch Von{\'a}sek, Daniel Fi{\v{s}}er, and
  Jan Faigl.
\newblock {\em Research and Education in Robotics - EUROBOT 2011: International
  Conference, Prague, Czech Republic, June 15-17, 2011. Proceedings}.
\newblock Springer Berlin Heidelberg, Berlin, Heidelberg, 2011.

\bibitem[\protect\citeauthoryear{Roberts \bgroup \em et al.\egroup
  }{2002}]{roberts2002low}
Jonathan~M Roberts, Peter~I Corke, and Gregg Buskey.
\newblock Low-cost flight control system for a small autonomous helicopter.
\newblock In {\em Australasian Conference on Robotics and Automation}.
  Auckland, New Zealand, 2002.

\bibitem[\protect\citeauthoryear{Ross \bgroup \em et al.\egroup
  }{2013}]{ross2013learning}
Susan Ross, Narek Melik-Barkhudarov, Kumar~Shaurya Shankar, Andreas Wendel,
  Debabrata Dey, J~Andrew Bagnell, and Martial Hebert.
\newblock Learning monocular reactive uav control in cluttered natural
  environments.
\newblock In {\em Proceedings of ICRA 2013}, pages 1765--1772. IEEE, 2013.

\bibitem[\protect\citeauthoryear{Shakernia \bgroup \em et al.\egroup
  }{1999}]{shakernia1999landing}
Omid Shakernia, Yi~Ma, T~John Koo, and Shankar Sastry.
\newblock Landing an unmanned air vehicle: Vision based motion estimation and
  nonlinear control.
\newblock {\em Asian journal of control}, 1(3):128--145, 1999.

\bibitem[\protect\citeauthoryear{Sharp \bgroup \em et al.\egroup
  }{2001}]{sharp2001vision}
Courtney~S Sharp, Omid Shakernia, and S~Shankar Sastry.
\newblock A vision system for landing an unmanned aerial vehicle.
\newblock In {\em Proceedings of ICRA 2001}, volume~2, pages 1720--1727. IEEE,
  2001.

\bibitem[\protect\citeauthoryear{Trisiripisal \bgroup \em et al.\egroup
  }{2006}]{trisiripisal2006stereo}
Phichet Trisiripisal, Matthew~R Parks, A~Lynn Abbott, Tianshu Liu, and Gary~A
  Fleming.
\newblock Stereo analysis for vision-based guidance and control of aircraft
  landing.
\newblock In {\em In 44th AIAA Aerospace Science Meeting and Exhibit (pp.
  1-17).}, 2006.

\bibitem[\protect\citeauthoryear{Veloso \bgroup \em et al.\egroup
  }{2012}]{6386300}
M.~Veloso, J.~Biswas, B.~Coltin, S.~Rosenthal, T.~Kollar, C.~Mericli,
  M.~Samadi, S.~Brandão, and R.~Ventura.
\newblock Cobots: Collaborative robots servicing multi-floor buildings.
\newblock In {\em Proceedings of IROS 2012}, pages 5446--5447, Oct 2012.

\bibitem[\protect\citeauthoryear{Ventura \bgroup \em et al.\egroup
  }{2013}]{ventura2013web}
Rodrigo Ventura, Brian Coltin, and Manuela Veloso.
\newblock Web-based remote assistance to overcome robot perceptual limitations.
\newblock In {\em AAAI-13, Workshop on Intelligent Robot Systems.}, 2013.

\bibitem[\protect\citeauthoryear{Webb \bgroup \em et al.\egroup
  }{2007}]{webb2007vision}
Thomas~P Webb, Richard~J Prazenica, Andrew~J Kurdila, and Rick Lind.
\newblock Vision-based state estimation for autonomous micro air vehicles.
\newblock {\em Journal of guidance, control, and dynamics}, 30(3):816--826,
  2007.

\bibitem[\protect\citeauthoryear{Wu \bgroup \em et al.\egroup
  }{2005}]{wu2005vision}
Allen~D Wu, Eric~N Johnson, and Alison~A Proctor.
\newblock Vision-aided inertial navigation for flight control.
\newblock {\em Journal of Aerospace Computing, Information, and Communication},
  2(9):348--360, 2005.

\bibitem[\protect\citeauthoryear{Yakimenko \bgroup \em et al.\egroup
  }{2002}]{yakimenko2002unmanned}
Oleg~A Yakimenko, Isaac~I Kaminer, William~J Lentz, and PA~Ghyzel.
\newblock Unmanned aircraft navigation for shipboard landing using infrared
  vision.
\newblock Technical report, DTIC Document, 2002.

\bibitem[\protect\citeauthoryear{Zhang and Ostrowski}{1999}]{zhang1999visual}
Hong Zhang and James~P Ostrowski.
\newblock Visual servoing with dynamics: Control of an unmanned blimp.
\newblock In {\em Robotics and Automation, 1999. Proceedings. 1999 IEEE
  International Conference on}, volume~1, pages 618--623. IEEE, 1999.

\end{thebibliography}


\end{document}